\documentclass{article}
\usepackage{spconf,amsmath,graphicx}
\usepackage{multirow}
\usepackage{booktabs}  
\usepackage{threeparttable}  

\title{Cascaded Residual Density Network for Crowd Counting}
\name{Kun Zhao, Luchuan Song, Bin Liu, Qi Chu and Nenghai Yu}
\address{School of Information Science and Technology, University of Science and Technology of China\\Key Laboratory of Electromagnetic Space Information, the Chinese Academy of Sciences \\ kunzhao@mail.ustc.edu.cn, flowice@ustc.edu.cn, slc0826@mail.ustc.edu.cn, $\{$whli, ynh$\}$@ustc.edu.cn}
\begin{document}
%
\maketitle
\begin{abstract}
Crowd counting is a challenging task due to the issues such as scale variation and perspective variation in real crowd scenes. In this paper, we propose a novel Cascaded Residual Density Network (CRDNet) in a coarse-to-fine approach to generate the high-quality density map for crowd counting more accurately. (1) We estimate the residual density maps by multi-scale pyramidal features through cascaded residual density modules. It can improve the quality of density map layer by layer effectively. (2) A novel additional local count loss is presented to refine the accuracy of crowd counting, which reduces the errors of pixel-wise Euclidean loss by restricting the number of people in the local crowd areas. Experiments on two public benchmark datasets show that the proposed method achieves effective improvement compared with the state-of-the-art methods.
\end{abstract}
\begin{keywords}
Crowd counting, Scale variation, CRDNet, Local count loss
\end{keywords}
\section{Introduction}
\label{sec:intro}

\begin{figure*}[ht]
  \centering
  \includegraphics[height=5.4cm]{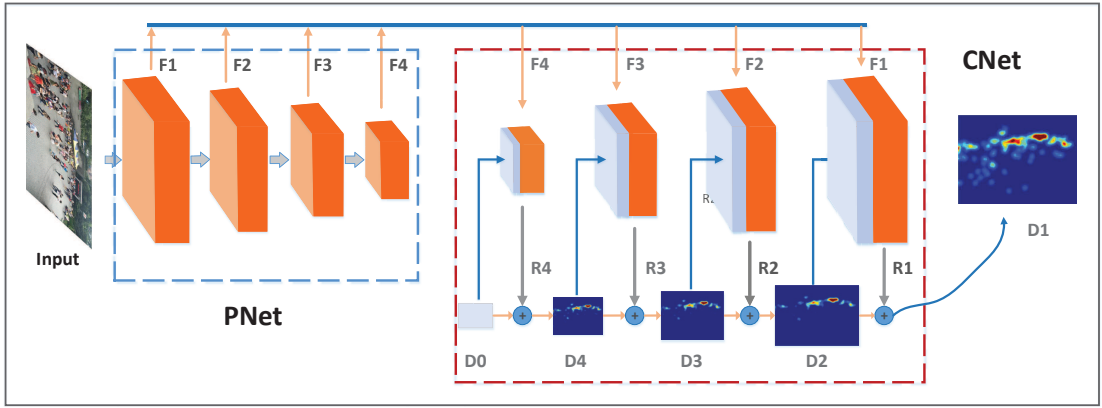}
%
\caption{The network structure of CDRNet, which consists of the encoder (PNet) and the decoder (CNet). PNet generates the pyramid of multi-scale features. And CNet estimates the density map through the cascaded residual density modules.}
\label{fig:network}
\end{figure*}

With the extensive use of urban monitoring systems, crowd counting is becoming increasingly important in the field of public security and attracts a lot of attention in recent years. Estimating the number of crowds accurately can effectively solve the problems of crowd control, crowd gathering monitoring and overload warning in practical applications. Furthermore, the method of crowd density estimation and counting can be extended to other fields like vehicle counting, urban building density estimation, etc. However, it is still a challenging work to generate the high-quality crowd density map for crowd counting. The major obstacle comes from the large scale variation in surveillance videos.

Recent works based on Convolutional Neural Networks utilize multi-scale architectures to address the problem of scale variation and have made great progress in crowd density estimation. For example, the multi-column networks~\cite{zhang2016single,sam2017switching,sindagi2017generating} were proposed by using multiple parallel columns with different filter sizes (large, medium, small). In such networks, the different column CNN learned different scale features to adaptive to a large variation in people or head size due to the camera perspective. Besides, the Multi-scale Network~\cite{zeng2017multi} based on the Inception-like Module was used to extract the scale-relevant features which could generate the crowd density map with various scales.

However, the above methods suffer inherent disadvantages. On the one hand, those algorithms utilize different filter sizes to extract corresponding features at different scales, but the performance is restricted by the types of filter. It may result in a lower quality of the estimated density map. On the other hand, most of the algorithms adopt pixel-wise Euclidean loss to optimize networks. As we know, the Euclidean loss is sensitive to outliers and blur the generated image~\cite{isola2017image}, which may affect the accuracy of crowd counting.

In this paper, to deal with these issues, we propose a novel crowd counting network named Cascade Residual Density Network (CRDNet), which adopts U-Net architecture~\cite{ronneberger2015u} that consists of an encoder and a decoder. Unlike previous approaches, where multiple filters with different kernel sizes are needed to extract multi-scale features for estimating crowd density map, the proposed CRDNet approach utilizes pyramidal feature extraction inspired by~\cite{hui2018liteflownet} in the encoder, which has more scale context information including strong semantic features and weak semantic features at different levels from a single input image scale. More importantly, in the proposed decoder, the high-quality crowd density map is estimated by summing up the residual density map in a coarse-to-fine framework. At each level of the pyramid, the proposed residual density module learns the residual density map for correcting the estimated density map of the previous level. Firstly, the stacked residual modules integrate the features at all scales naturally, which can be more robust to scale variation in the task of crowd counting. Secondly, the cascaded structure progressively improves the resolution and quality of density maps by estimating the residual density maps without passing more errors to the next level. Therefore the proposed architecture estimates the crowd density map layer by layer that can improve the accuracy of crowd counting effectively.

Furthermore, considering that the Euclidean loss only focuses on the errors of individual pixels, it is usually affected by outlier pixels. In this paper, an additional local count loss, which is defined as the errors of crowd counting between the estimated density map and the ground truth in local areas, is proposed to describe the accuracy of the total number of the local crowd rather than the single pixel. It can regularize the estimated density map and reduce the counting error caused by the Euclidean loss. The experiment results demonstrate that it can greatly improve the performance. In summary, the key highlights of the work are:

(1) The cascaded density residual network is proposed to generate the high-quality crowd density map from coarse to fine through stacked residual density modules in a pyramid of multi-scale features. 
 
(2) An additional local count loss is designed to minimize the errors of local crowd counting not only the pixel value, which constrains the Euclidean distance between estimated density map and the ground truth pixels. It improves the whole quality of estimated crowd density maps effectively.

\section{Related works}
\label{sec:format}

In recent years, various approaches have been proposed in the literature for crowd counting. Early traditional crowd counting methods that use hand-crafted features can be classified into the two categories: (1) Detection-based methods, (2) Regression-based methods. 

In detection-based methods, the number of crowds is calculated by detecting the individual entities in the scene~\cite{dalal2005histograms,leibe2005pedestrian,enzweiler2009monocular,zhao2008segmentation}. Although such approaches have achieved good results in low-density scenes, the problem of occlusion among people in the high-density crowds adversely affects the performance of accurate detection. In regression-based methods, researchers avoid detecting the people in the scene but turn the crowd counting into regression problem by learning a mapping between holistic or local features extracted from images to their count~\cite{chan2009bayesian,ryan2009crowd,chen2012feature}. The regression techniques include linear regression, ridge regression, Gaussian regression etc.
		
More recently, a variety of CNN-based methods for crowd counting have achieved significant improvement and outperformed traditional methods with handcrafted features due to the powerful feature characterization of CNN. Zhang \textit{et al}.~\cite{zhang2015cross} trained a CNN to regress the crowd density map, and fine-tuned the trained network to adapt to a new scene by retrieving image samples that were similar to the new scene. But this method requires perspective maps to solve the large scale and perspective variations in a crowd scene, which limits many practical applications. Zhang \textit{et al}.~\cite{zhang2016single} proposed a multi-column network (MCNN) to extract features for crowd counting without perspective maps. The network of different columns corresponding to different filter sizes (large, medium, small) ensured the robustness to large variation in object scales. Instead of training a multi-column network on all the input images, Sam \textit{et al}.~\cite{sam2017switching} proposed an extra switching CNN that was trained to select an independent CNN regressor suited for a given input. The independent CNN regressors were designed with different filter sizes for different scales. Similarly, Onoro \textit{et al}.~\cite{onoro2016towards} proposed a scale-aware crowd counting network named Hydra CNN that learned a multi-scale non-linear regression model by using a pyramid of image patches extracted at multiple scales. In addition, Li \textit{et al}~\cite{li2018csrnet} utilized dilated kernels to deliver larger reception fields for crowd counting tasks.  Ranjan \textit{et al}~\cite{ranjan2018iterative} tackled the problem of crowd counting by combing low-resolution CNN branch and high-resolution CNN branch respectively.

\section{Cascaded Residual Density Network}
\label{sec:pagestyle}

The Cascade Residual Density Network consists of pyramidal feature extraction and Cascade density map estimation, named PNet and CNet respectively as shown in Fig.~\ref{fig:network}. The PNet extracts pyramids of multi-scale features from the different layers of the backbone network to represent the scale diversity of features. And the CNet consists of cascaded residual density modules that learn coarse-to-fine density map for crowd counting.

\subsection{Pyramidal Feature Extraction}
\label{sssec:subsubhead}

As shown in Fig.~\ref{fig:network}, PNet is a feature extractor, which utilizes the CSRnet~\cite{li2018csrnet} based on VGG16~\cite{Simonyan14c} as the backbone. We extract the multi-scale pyramidal features $F_k$ from the mid layers of PNet ($conv1\_2$, $conv2\_2$, $conv3\_3$, $conv5\_6$), where $k$ denotes the level of pyramid. In this paper, we construct a four-layer feature pyramid to adapt to the scale variances in crowd scenes. The resolutions of pyramidal features reduce by a factor $s$ from top ($k=1$) to bottom ($k=4$) due to the operations of pooling and stride convolutions.

\subsection{Cascaded Density Map Estimation}
\label{sssec:subsubhead}
CNet is composed of multiple residual density modules as shown in Fig.~\ref{fig:network}. At each pyramid level of PNet, the residual density map is estimated by the corresponding residual density module, and the cascaded structure further improves the quality of density map from low-resolution to high-resolution. 
\vspace{-0.3cm}
\begin{figure}[htb]
  \centering
  \includegraphics[height=3.2cm]{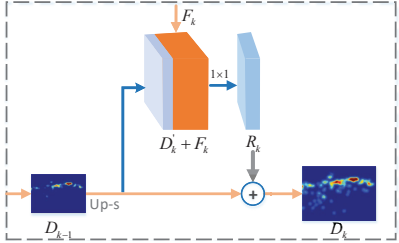}
\caption{The residual density module $G_k$ }
\label{fig:netw}
\end{figure}

As shown in Fig.~\ref{fig:netw}, at the $k$-th level of pyramid, the residual density module $G_k$ transforms the upsampled density map ${D'_k}$ from the previous level $D_{k-1}$ and the $k$-th level pyramidal features $F_k$ to the residual density map $R_k$. $D_0$ represents the crowd density map initialized to zero. The calculation process is as follows:

\begin{equation}
\label{equ:res}
R_k = C_k({f(u_s(D_{k-1}),F_k)})
\end{equation} 
where~${u_s}( \cdot )$ denotes the function of upsampling by a factor $s$ using bilinear interpolation. $f(\cdot)$ represents that features are concatenated in the channel dimension. $C_k$ denotes the convolution operation with the filter of size $1 \times 1$.

Then, the density map $D_k$ at the $k$-th pyramid level can be represented as:

\begin{equation}
\label{equ:density}
D_k = u_s(D_{k-1}) + R_k
\end{equation}

The estimated $D_k$ is similarly passed on to the next higher resolution level until we generate the final density map $D_1$ with the same size as the input image. In summary, the final estimated density map $D_1$ is ceaselessly refined by adding residual density maps of various scales in the cascaded network, which enhances the quality of crowd density maps.

\subsection{Crowd Counting Loss}
\label{sssec:subsubhead}

\textbf{Euclidean Loss.} The pixel-wise Euclidean loss is used to measure the distance between the estimated density map and the ground truth in most works for crowd counting. The Euclidean loss function~$L_E(\Theta )$ is defined as follows:

\begin{equation}
\label{equ:2}
L_E(\Theta ) = {1 \over M}\sum\limits_{j = 1}^M {{{\left\| {{D_Q}({m_j};\Theta ) - {Q_j}} \right\|}^2}} 
\end{equation}
where $\Theta$ is the set of parameters of the CRDNet model.~$M$ denotes the number of training patch images.~$m_j$ is the input image of the network. $D_Q$ denotes the estimated density map, and $Q_j$ is the corresponding ground truth density map.  
\begin{figure}[htb]
  \centering
  \includegraphics[height=2.8cm]{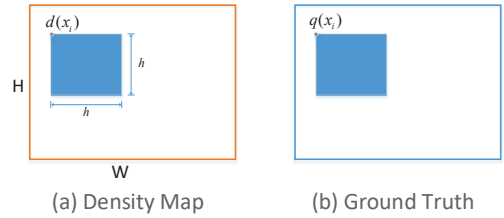}
%
\caption{The estimated density map $D_Q$ (a) and the ground truth $Q$ (b)}
\label{fig:netlocal}
\end{figure}
\setlength\parskip{.5\baselineskip}

\textbf{Local Count Loss}. The proposed local count loss is used to calculate the difference of crowd counting between the estimated density map and the ground truth in local areas of the input image. The loss measures the number of people in the local crowd rather than a single pixel, which can optimize the network and pay more attention to the details of local areas on the basis of Euclidean loss. Then it can be more beneficial to the overall crowd counting. The local count loss ($L_Y(\Theta)$ is defined as follows:  
\begin{equation}
\label{equ:2}
{c_j}(\textbf{x}_i|\Theta) = \sum\limits_{\sigma  \in \left[ {0,h} \right] \times \left[ {0,h} \right]} {d_j({{\textbf{x}_i+\sigma;\Theta}})-q_j({{\textbf{x}_i+\sigma}})}  
\end{equation}
\begin{equation}
\label{equ:2}
{L_Y(\Theta)} = {1 \over M}\sum\limits_{j = 1}^M{\sum\limits_{i \in \left[ {0,H} \right] \times \left[ {0,W} \right], stride=t } \left|c_j(\textbf{x}_i|\Theta)\right| }
\end{equation}
where $d$ and $q$ denote the local patch of estimated density map $D_Q$ and the ground truth $Q$ respectively as shown in Fig.~\ref{fig:netlocal}. For each square patch of size $h$ whose upper left coordinate at $\textbf{x}_i$, we calculate the absolute count error $\left|c_j(\textbf{x}_i|\Theta)\right|$ for the $j$-th training image with no padding. $t$ represents the stride of the patch. And the total local count loss~$L_Y(\Theta)$ is the sum of all patch losses.

\setlength\parskip{.5\baselineskip}
\textbf{Total Loss}. We combine the Euclidean loss and the local count loss as the final crowd counting loss in this paper, and it is defined as follows:
\begin{equation}
\label{equ:2}
L(\Theta) = L_E(\Theta) + \lambda {L_Y}(\Theta)
\end{equation}
where $\lambda$ denotes the weight of $L_Y(\Theta)$. In our experiments, we empirically set $\lambda$ as 0.0001.

\section{Experiments}
\label{sec:exp}

In the training stage, firstly we train each residual density module $G_k$ independently to estimate the corresponding crowd density map at the $k$-th layer of the pyramid. The target residual density map $\tilde R_k$ is the difference target density map $H_k$ and the estimated density map in the previous layer. $\tilde R_k = H_k - u_s(D_{k-1})$. Finally, we fine-tune the whole network in the public datasets.

We evaluate the crowd counting methods with the absolute error (MAE) and mean square error (MSE), which are defined as follows:
\begin{equation}
\label{equ:mae}
MAE = {1 \over {N}}\sum\limits_{j = 1}^{N} {\left| {{n_j} - {{\hat n}_j}} \right|}, MSE = \sqrt {{1 \over {N}}\sum\limits_{j = 1}^{N} {{{({n_j} - {{\hat n}_j})}^2}} }
\end{equation}
where $N$ is the number of test images. For the $j$-th sample, $n_j$ denotes the ground truth count and $\hat n_j$ denotes the estimated crowd count that is the integral of the estimated density map.

The proposed method is tested on two public datasets including ShanghaiTech dataset and UCF$\_$CC$\_$50 dataset. The results show the effectiveness and the superiority compared with the state-of-the-art methods. 

\subsection{ShanghaiTech Dataset}

The ShanghaiTech dataset~\cite{zhang2016single} is divided into two parts: Part$\_$A and Part$\_$B. In the Part$\_$A sub-dataset, there are 482 images crawled from the internet, and in the Part$\_$B sub-dataset, there are 716 images collected from the busy street.

We compare our method with the existing state-of-the-art methods using MAE and MSE metrics. In Table~\ref{tab:shanghai}, it can be observed that the proposed CRDNet can achieve superior performance compared with those networks with multiple filter size~\cite{zhang2016single,sam2017switching,zeng2017multi}, which demonstrates that the pyramidal features have strong semantics at all scales. 
\vspace{-0.3cm}
\renewcommand{\arraystretch}{1.4} 
\begin{table}[!htb]  
  \caption{Comparison of CRDNet with other state-of-the-art methods on the ShanghaiTech dataset.} 
  \centering  
  \fontsize{8}{8}\selectfont  
  \begin{threeparttable}  
   
  \label{tab:shanghai}  
    \begin{tabular}{p{2.4cm}p{1cm}p{1cm}p{1cm}p{1cm}}  
    \toprule  
    \multirow{2}{*}{Method}&  
    \multicolumn{2}{c}{ Part$\_$A}&\multicolumn{2}{c}{ Part$\_$B}\cr  
    \cmidrule(lr){2-3} \cmidrule(lr){4-5} 
    &MAE&MSE&MAE&MSE\cr  
    \midrule  
		Zhang \textsl{et al}.~\cite{zhang2015cross}&181.8&277.7&32.0&49.8\cr  
    MCNN~\cite{zhang2016single}&110.2&173.2&26.4&41.3\cr 
    Switch-CNN~\cite{sam2017switching}&90.4&135.0&21.6&33.4\cr
		CP-CNN~\cite{sindagi2017generating}&73.6&106.4&20.1&30.1\cr
		MSCNN~\cite{zeng2017multi}&83.8&127.4&26.4&41.3\cr
    CSRNet~\cite{li2018csrnet}&{\bf 68.2}&115.0&10.6&16.0\cr 
		ic-CNN~\cite{ranjan2018iterative}&68.5&116.2&10.7&16.0\cr	
    ACSCP~\cite{shen2018crowd}&75.7&{\bf 102.7}&17.2&27.4\cr   
    \textbf{CRDNet (proposed)}&{ 68.5}&{ 108.4}&{\bf 8.5}&{\bf 13.6}\cr  
    \bottomrule  
    \end{tabular}  
    \end{threeparttable}  
\end{table} 

\subsection{UCF$\_$CC$\_$50 Dataset}
\vspace{-0.3cm}
The UCF$\_$CC$\_$50 dataset~\cite{idrees2013multi} consists of 50 images collected from the publicly available web. We perform 5-fold cross-validation to evaluate the effectiveness of our proposed CRDNet method.
As shown in Table~\ref{tab:ucf}, in the high-density crowd scenes, the proposed CRDNet generates high-quality density maps by the cascaded network, which makes great progress compared with the state-of-the-art methods in the MAE metric. In the MSE metric, it can also obtain a state-of-the-art score.
\vspace{-0.1cm}
\renewcommand{\arraystretch}{1.2} 
\begin{table}[!htb]
  \caption{Comparison of CRDNet with other state-of-the-art methods on the UCF$\_$CC$\_$50 dataset.} 
  \centering  
  \fontsize{7}{7}\selectfont  
  \begin{threeparttable}  
   
  \label{tab:ucf}  
    \begin{tabular}{p{2.5cm}p{1.5cm}p{1.2cm}}  
    \toprule  
		\multirow{1}{*}{Method}&
    MAE&MSE\cr  
    \midrule
    Zhang \textit{et al}.~\cite{zhang2015cross}&467.0&498.5\cr 
    MCNN~\cite{zhang2016single}&377.6&509.1\cr  
    Switch-CNN~\cite{sam2017switching}&318.1&439.2\cr
		CP-CNN~\cite{sindagi2017generating}&295.8&{\bf320.9}\cr
		MSCNN~\cite{zeng2017multi}&363.7&468.4\cr
    CSRNet~\cite{li2018csrnet}&{266.1}&397.5\cr 
		ic-CNN~\cite{ranjan2018iterative}&260.9&365.5\cr	
    ACSCP~\cite{shen2018crowd}&291.0&404.6\cr   
    \textbf{CRDNet (proposed)}&{\bf 250.7}&{ 346.6}\cr  
    \bottomrule  
    \end{tabular}  
    \end{threeparttable}  
\end{table}  
\vspace{-0.8cm}
\subsection{Loss Function}
\vspace{-0.2cm}
To evaluate the effectiveness of the local count loss, we compare the performance of different loss functions on public datasets as shown in Table~\ref{tab:ly}. The results demonstrate that $L_Y(\Theta)$ can increase the accuracy of crowd counting.
\vspace{-0.3cm}
\renewcommand{\arraystretch}{1.0} 
\begin{table}[!htb]  
  \caption{ Comparisons of errors with different losses on the public benchmark datasets.}  
  \centering  
  \fontsize{7}{7}\selectfont  
  \begin{threeparttable}  
  
  \label{tab:ly}  
    \begin{tabular}{ccccccc}  
    \toprule  
    \multirow{3}{*}{Loss Function}&  
    \multicolumn{2}{c}{ Part$\_$A}&\multicolumn{2}{c}{ Part$\_$B}&\multicolumn{2}{c}{ UCF$\_$CC$\_$50}\cr  
    \cmidrule(lr){2-3} \cmidrule(lr){4-5}\cmidrule(lr){6-7} 
    &MAE&MSE&MAE&MSE&MAE&MSE\cr  
    \midrule  
    $L_E$&70.3&110.2&8.5&13.7&255.6&363.1\cr   
    $L_E$, $L_Y$&{\bf 68.5}&{\bf 108.4}&{\bf 8.5}&{\bf 13.6}&{\bf 250.7}&{\bf 346.6}\cr  
    \bottomrule  
    \end{tabular}  
    \end{threeparttable}  
\end{table}  

\vspace{-0.4cm}

\section{conclusion}
\label{sec:con}
\vspace{-0.2cm}
In this paper, the proposed Cascaded Residual Density Network takes advantage of the feature pyramid structure to extract multi-layer feature maps at various scales and refines the density map by cascaded coarse-to-fine architecture. To better improve the accuracy of crowd counting, an additional local crowd loss is proposed to regularize the estimated density map. Extensive experiments performed on the ShanghaiTech and UCF$\_$CC$\_$50 datasets demonstrate the effectiveness compared with recent state-of-the-art methods. 

\noindent\textbf{Acknowledgement.}
This work is supported by the National Natural Science Foundation of China (Grant No. 61371192), the Key Laboratory Foundation of the Chinese Academy of Sciences (CXJJ-17S044) and the Fundamental Research Funds for the Central Universities(WK2100330002, WK3480000005).

\bibliographystyle{IEEEbib}
\bibliography{strings,refs}

\end{document}